# Unsupervised Deep Domain Adaptation for Pedestrian Detection

Lihang Liu[1], Weiyao Lin[1*], Lisheng Wu[1], Yong Yu[1], Michael Ying Yang[2]

[1]Shanghai Jiao Tong University, Shanghai, China (*corresponding author)
[2]ITC-EOS, University of Twente, The Netherlands

**Abstract.** This paper addresses the problem of unsupervised domain adaptation on the task of pedestrian detection in crowded scenes. First, we utilize an iterative algorithm to iteratively select and auto-annotate positive pedestrian samples with high confidence as the training samples for the target domain. Meanwhile, we also reuse negative samples from the source domain to compensate for the imbalance between the amount of positive samples and negative samples. Second, based on the deep network we also design an unsupervised regularizer to mitigate influence from data noise. More specifically, we transform the last fully connected layer into two sub-layers — an element-wise multiply layer and a sum layer, and add the unsupervised regularizer to further improve the domain adaptation accuracy. In experiments for pedestrian detection, the proposed method boosts the recall value by nearly 30% while the precision stays almost the same. Furthermore, we perform our method on standard domain adaptation benchmarks on both supervised and unsupervised settings and also achieve state-of-the-art results.

**Keywords:** Unsupervised Domain Adaptation, Unsupervised Regularizer, Deep Neural Network, People Detection

## 1 Introduction

Deep neural networks have shown great power on traditional computer vision tasks, however, the labelled dataset should be large enough to train a reliable deep model. The annotation process for the task of pedestrian detection in crowded scenes is even more resource consuming, because we need to label concrete locations of pedestrian instances. In modern society, there are over millions of cameras deployed for surveillance. However, these surveillance situations vary in lights, background, viewpoints, camera resolutions and so on. Directly utilizing models trained on old scenes will result in poor performance on new situations due to data distribution changes. It is also unpractical to annotate pedestrian instances for every surveillance situation.

When there are few or no labelled data in the target domain, domain adaptation helps to reduce the amount of labelled data needed. Basically, unsupervised domain adaptation aims to shift the model trained from the source domain to the target domain for which only unlabelled data are provided. Most traditional works [1–5] either learn a shared representation between the source and target domain, or project features into a common subspace. Recently, there are also works [6–8] proposed to learn a scene-specific detector by deep architectures. However, heuristic methods are needed either for constructing feature space or re-weighting samples. Our motivation of developing a domain adaptation architecture is to reduce heuristic methods required during the adaptation process.

In this paper, we propose a new approach for unsupervised deep domain adaptation for pedestrian detection. First, we utilize an iterative algorithm to iteratively auto-annotate target examples with high confidence as positive pedestrian instances on the target domain. During each iteration, these auto-annotated data are regarded as the training set to update the target model. However, these auto-annotated samples still have the limitations of lack of negative samples and existence of false positive samples, which will no doubt lead to exploration of predictions on non-pedestrian instances. Therefore, in order to compensate for the quantitative imbalance between positive and negative samples, we randomly sample negative instances from the source domain and mix into training set. Second, based on deep network, we further design an unsupervised regularizer to mitigate influence from data noise and avoid overfitting. More specifically, in order to have a better regularization effect during the adaptation process, we propose to transform the last fully connected layer of the deep model into two sub-layers, an element-wise multiply layer and a sum layer. Thus, the unsupervised regularizer can be added on the element-wise multiply layer to adjust all weights in the deep network and gain better performance.

The contributions of our work are three folds.

- We propose an adaptation framework to learn scene-specific deep detectors for target domains by unsupervised methodologies, which adaptively selects positive instances with high confidence. This can be easily deployed to various surveillance situations without any additional annotations.
- Under this framework, we combine both supervised term and unsupervised regularizer into our loss function. The unsupervised regularizer helps to reduce influence from data noise in the auto-annotated data.
- More importantly, for better performance of the unsupervised regularizer we propose to transform the last fully connected layer of the deep network into two sub-layers, an element-wise multiply layer and a sum layer. Thus, all weights contained in the deep network can be adjusted under the unsupervised regularizer. To the best of our knowledge, this is the first attempt to transform fully connected layers for the purpose of domain adaptation.

The remainder of this paper is organized as follows. Section 2 reviews related works. Section 3 presents the details of our approach. Experimental results are shown in Section 4. Section 5 concludes the paper.

## 2  Related Work

In many detection works, the generic model trained from large amount of samples on the source domain is directly utilized to detect on the target domain. They assume that samples on the target domain are subsets of the source domain. However, when the distribution of data on the target and source domain vary largely, the performance will drop significantly. Domain adaptation aims to reduce the amount of data needed for the target domain.

Many domain adaptation works try to learn a common representation space shared between the source and target domain. Saenko et al. [1, 2] propose both linear-transform-based techniques and kernel-transform-based techniques to minimize domain changes. Gopalan et al. [3] project features into Grassmann manifold instead of operating on features of raw data. Alternatively, Mesnil et al. [9] use transfer learning to obtain good representations. However, these methods have limitations since scene-specific features are not learned to boost accuracy.

Another group of works [4, 5, 10, 11] on domain adaptation is to make the distribution of the source and target domain more similar. Among these works, Maximum Mean Discrepancy (MMD)[12] is used to as a metric to reselect samples from the source domain in order to have similar distribution as target samples. In [13], MMD is added on the last feature vector of the network as a regularization. Different from these methods, our work transforms the last fully connected layer into two sub-layers, an element-wise multiply layer and a sum layer. As the element-wise multiply layer is the last layer that contains weights before output layers, our unsupervised regularizer on the element-wise multiply layer can adjust all weights of the deep network during training.

There are also works on deep adaptation to construct scene-specific detectors. Wang et al.[6] explore context cues to compute confidence, [7] learn distributions of target samples and propose a cluster layer for scene-specific visual patterns. These works re-weight auto-annotated samples for their final object function and additional context cues are needed for reliable performance. However, heuristic methods are required to select reliable samples. Alternately, Hattori el al. [8] learn scene-specific detector by generating a spatially-varying pedestrian appearance model. And Pishchulin et al. [14] use 3D shape models to generate training data. However, synthesis for domain adaptation is also costly. Compared with these methods, our approach does not include the heuristic pre-processing steps. Thus, the performances of our approach are not affected by the pre-processing steps.

## 3  Our Approach

In this section, we introduce our unsupervised domain adaptation architecture on the task of pedestrian detection in crowded scenes. Unsupervised domain adaptation aims to shift the model trained from the source domain to the target domain for which only unlabelled data are provided. Under the unsupervised setting, we use an iterative algorithm to iteratively auto-annotate target samples and update the target model. As the auto-annotated samples may contain noises, the performances may be affected by the wrongly annotated samples. Therefore, an unsupervised regularizer is introduced to mitigate the influence from data noise on the target model. More specifically, based on the assumption that the source domain and the target domain should share the same feature space after feature extraction layers, we encode the unsupervised regularizer to make a constraint that the distribution of data representation on the element-wise multiply layer should be similar between the source domain and the target domain.

The adaptation architecture of our approach consists of three parts – the source stream, the target stream and an unsupervised regularizer, as shown in Fig 1. The source stream takes samples from the source domain as input, while the target stream is trained from auto-annotated positive samples from the target domain and negative samples from the source domain. These two streams can utilize any deep detection network as their basic model, as well as their detection loss function as supervised loss functions of two streams. In our experiments, we use the detection network mentioned in Section 4.1 as the basic model. The unsupervised regularizer is integrated into the loss function of the target stream.

In the following, we will first describe our iterative algorithm which iteratively selects samples from the target domain, and updates the target model accordingly (Section 3.1). Then, we will introduce the loss function we designed for updating the target model (Section 3.2), as well as the proposed unsupervised regularizer for improving the domain adaptation performance (Section 3.3).

### 3.1   Iterative Algorithm

In this section, we introduce the iterative algorithm which is the training method of the target stream of our adaptation architecture. There are two reasons to employ the iterative algorithm. First, auto-annotated data on the target domain vary for every adaptation iteration and new positive samples will be auto-annotated as training set. Compared to methods without the iterative algorithm, it helps to avoid overfitting caused by lack of data. Second, unsupervised regularizer performs better with more training data as it's a distribution based regularizer.

There are two stages for the iterative algorithm. The source stream and the target stream are separately trained at different stages. At initialization stage, the source model of the source stream are trained under a supervised loss function with abundant labelled data, $(\mathbf{X}^S,\mathbf{Y}^S)$, from the source domain. After its convergence, the weights of the source model $\theta^S$ are taken to initialize the target stream. At adaptation stage, the target model is trained from auto-annotated positive samples $(\mathbf{X}^{T,n},\mathbf{Y}^{T,n})$ from the target domain and randomly-selected negative samples $(\mathbf{X}^{S,n},\mathbf{Y}^{S,n})$ from the source domain under both supervised loss function and unsupervised regularizer. Since auto-annotated data are all regarded as positive samples, negative samples from the source domain are randomly selected to compensate for lack of negative instances, which are human annotated and can thus provide true negative samples. Note that we do not jointly train two streams at adaptation stage and the weights of the source model stay static which serves as a distribution reference for the unsupervised regularizer at the adaptation stage. The complete adaptation process is illustrated in Algorithm 1. After a predetermined iteration limit $N^I$ is reached, we obtain our final detection model on the target domain.

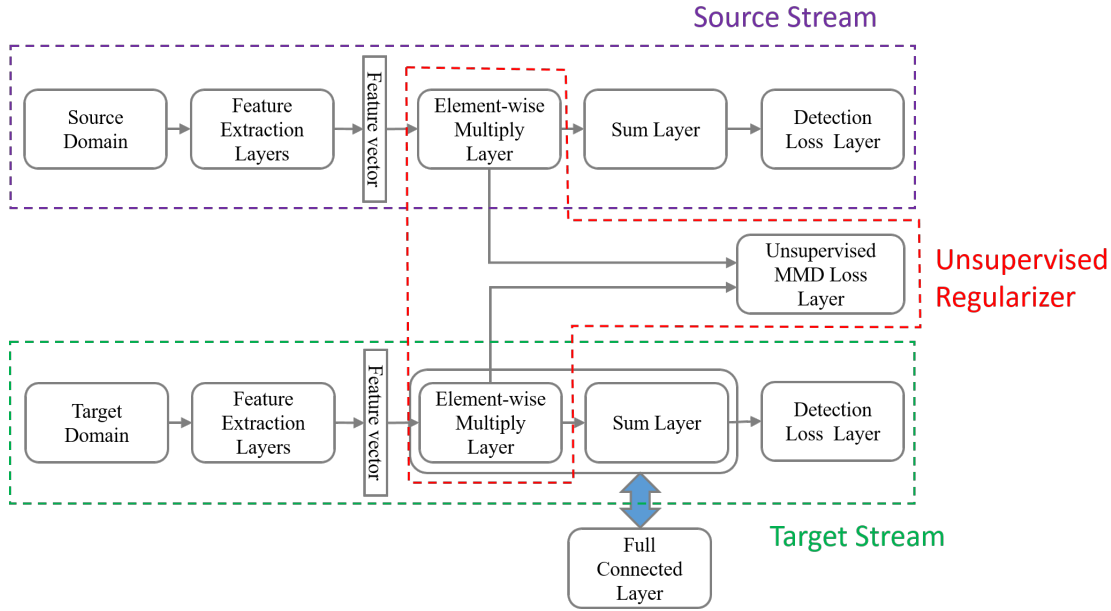

**Fig. 1.** The adaptation architecture consists of three parts, the source stream, the target stream and an unsupervised regularizer. The last fully connected layers of both source and the target stream are transformed into element-wise layer and sum layer for the purpose of the unsupervised regularizer. Best view in colors.

### 3.2 Loss function for the target stream

In this section, we introduce our loss function on the target stream of our adaptation architecture, which is composed of a supervised loss and an unsupervised regularizer. The supervised loss is to learn the scene-specific bias for the target domain, while the unsupervised regularizer introduced in Section 3.3 plays an important part in reducing influence from data noise as well as avoiding overfitting.

We denote training samples from the source domain as $\mathbf{X}^S = \{x_i^S\}_{i=1}^{N^S}$. For training samples on the source domain, we have corresponding annotations $\mathbf{Y}^S = \{y_i^S\}_{i=1}^{N^S}$ with $y_i^S = (b_i^S, l_i^S)$, where $b_i^S = (x, y, w, h) \in R^4$ is the bounding box location and $l_i^S \in \{0, 1\}$ is the label indicating whether $x_i^S$ is a pedestrian instance. At the $n^{th}$ adaptation iteration, we have two set of training samples, $N^{T,n}$ auto-annotated positive samples from the target domain $\mathbf{X}^{T,n} = \{x_j^{T,n}\}_{j=1}^{N^{T,n}}$ and $N^{T,n}$ negative samples from the source domain $\mathbf{X}^{S,n} = \{x_k^{S,n}\}_{k=1}^{N^{T,n}}$. Their corresponding annotations can be denoted as $\mathbf{Y}^{T,n} = \{y_j^{T,n}\}_{j=1}^{N^{T,n}}$ and $\mathbf{Y}^{S,n} =$

**Algorithm 1** Deep domain adaptation algorithm

1: **procedure** DEEP DOMAIN ADAPTATION
2: Train the source model $M^S$ on the source stream with abundant annotated data $(\mathbf{X}^S, \mathbf{Y}^S)$
3: Use $M^S$ to initialize the target model on the target stream as $M_0$
4:    **for** i = 0:$N^I$ **do**
5:       $M_i$ generate auto-annotated positive samples $(\mathbf{X}^{T,n}, \mathbf{Y}^{T,n})$ of the target domain
6:       Randomly sampled negative instances $(\mathbf{X}^{S,n}, \mathbf{Y}^{S,n})$ from the source domain
7:       $\mathbf{X}^n = \{\mathbf{X}^{T,n}, \mathbf{X}^{S,n}\}$
8:       $\mathbf{Y}^n = \{\mathbf{Y}^{T,n}, \mathbf{Y}^{S,n}\}$
9:       Take $(\mathbf{X}^n, \mathbf{Y}^n)$ as training data to upgrade $M_i$ into $M_{i+1}$
10:    **end for**
11: $M_{N^I}$: final model.
12: **end procedure**

$\{y_k^{S,n}\}_{k=1}^{N^{T,n}}$ with $y_j^{T,n} = (b_j^{T,n}, l_j^{T,n} \equiv 1, c_j^{T,n})$, and $y_k^{S,n} = (b_k^{S,n}, l_k^{S,n} \equiv 0)$, respectively. $c_*^{T,n}$ is the confidence given by the auto-annotation tool and $N^I$ is the maximum number of adaptation iterations. Now we can formulate the combination of supervised loss and unsupervised regularizer as follows:

$$L(\theta^{T,n}|\mathbf{X}^{T,n}, \mathbf{Y}^{T,n}, \mathbf{X}^{S,n}, \mathbf{Y}^{S,n}, \mathbf{X}^S, \theta^S) = L_S + \alpha * L_U \tag{1}$$

$$L_S = \sum_{j=1}^{N^{T,n}} H(c_j^{T,n}) * (R(\theta^{T,n}|x_j^{T,n}, b_j^{T,n}) + C(\theta^{T,n}|x_j^{T,n}, l_j^{T,n}))$$

$$+ \sum_{k=1}^{N^{T,n}} (R(\theta^{T,n}|x_k^{S,n}, b_k^{S,n}) + C(\theta^{T,n}|x_k^{S,n}, l_k^{S,n})) \tag{2}$$

$$L_U = L_{EWM}(\theta^{T,n}|\mathbf{X}^T, \mathbf{X}^S, \theta^S) \tag{3}$$

where $L_S$ is the supervised loss to learn scene-specific detectors and $L_U$ is the unsupervised regularizer part. $\alpha = 0.8$ is the coefficient balancing the effect of supervised and unsupervised loss. $\theta^{T,n}$ denote the coefficients of the network in the target stream at $n^{th}$ adaption and $\theta^S$ denote the coefficients of the network in the source stream. $H(\cdot)$ is a step function in order to select positive samples with high confidence among auto-annotated data on the target domain. $R(\cdot)$ is a regression loss for bounding box locations, such as norm-1 loss, and $C(\cdot)$ is a classification loss for bounding box confidence, such as cross-entropy loss. And $L_{EWM}(\cdot)$, to be introduced in Section 3.3, is a MMD-based loss added on the element-wise multiply layer for unsupervised regularization.

### 3.3   Unsupervised weights regularizer on Element-wise Multiply Layer

As mentioned before, the unsupervised regularizer plays an important role in reducing influence from data noise and avoiding overfitting. In this paper, we propose to transform the last fully connected layer in order to have better effect on unsupervised regularization.

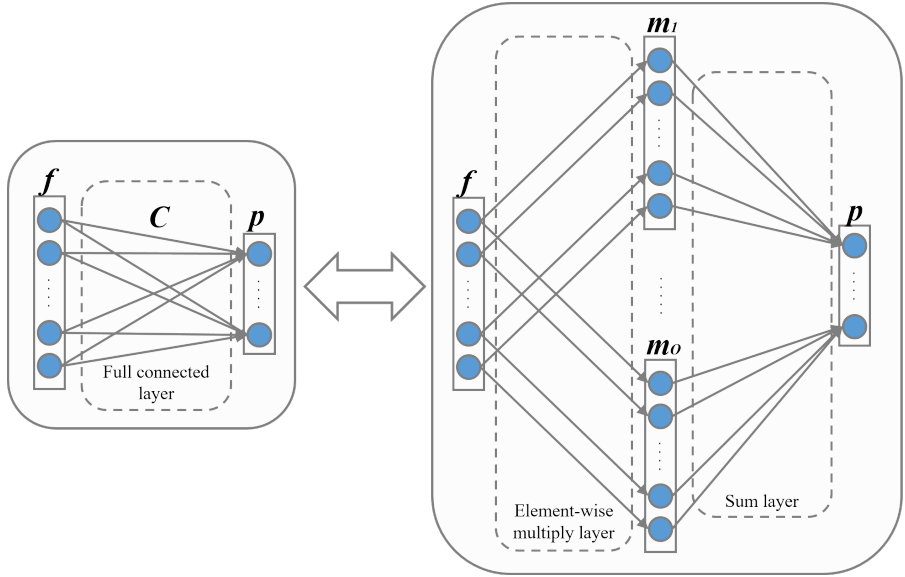

**Fig. 2.** Illustration of transformation of the fully connected layer **C** into element-wise multiply layer and sum layer. After the transformation, the element-wise layer become the last layer which contains weights before output layer $\boldsymbol{p}$. Thus, an unsupervised regularizer can be added on $\boldsymbol{m_o}$.

**Element-wise Multiply Layer** In deep neural network, the data of the last feature vector layer is taken as an important data representation of input images. However, in this paper, we take one step further to focus on the last fully connected layer which serves as an decoder to decode rich information of the last feature vector into final outputs. As the source model is trained with abundant labelled data on the source domain, the weights of the last fully connected layer are also well converged. A regularizer on the last fully connected layer can adjust all weights of the network compared with that on the last feature vector layer. Denote the last feature vector, the weights of the last fully connected layer and the final outputs as $\boldsymbol{f}_{(1 \times N^D)}$, $\mathbf{C}_{(N^D \times N^O)}$ and $\boldsymbol{p}_{(1 \times N^O)}$. $N^D, N^O$ are the dimension of feature vector and the dimension of output layer, respectively. Thus the operation of the fully connected layer can be formulated as matrix multiply:

$$\boldsymbol{p} = \boldsymbol{f} * \mathbf{C} \qquad (4)$$

where

$$p_o = \sum_d f_d * C_{d,o} \qquad (5)$$

Inspired by this form, we separate the above formula into two sub-operations – the element-wise multiply operation and the sum operation, which can be

formulated as:

$$\boldsymbol{m}_o = [f_d * C_{d,o}]_{d=1}^{N^D} \tag{6}$$

$$\boldsymbol{p}_o = \boldsymbol{m}_o * \overrightarrow{\mathbf{1}} \tag{7}$$

where $\mathbf{M}_{(N^O \times N^D)} = [\boldsymbol{m}_o]$ is the intermediate results of the element-wise multiply operation. $\boldsymbol{m}_o$ is a vector with $N^D$ dimensions, which will be the object of the unsupervised regularizer. Finally, we can equivalent-transform the last fully connected layer into an element-wise multiply layer and a sum layer. The transformed element-wise multiply layer is thus the last layer with weights before output layers. Fig 2 illustrates the transformation.

**Unsupervised regularizer on Element-wise Multiply Layer** This section introduces our unsupervised regularizer. As stated in Section 3.1, there are false positive samples among auto-annotated data, which will mislead the network and result in worse performance. Thus, we designed an unsupervised regularizer to mitigate the influence. We have the assumption that the weights of the element-wise multiply layer of the last fully connected layer have well converged under the training of abundant source samples. Thus, when tasks are similar, the distribution of data representations of the element-wise multiply layer on the source domain and the target domain should also be similar. While false samples are easier to mutate the distribution of data representations. This observation can be illustrated in Fig 3, where the center of $m_o$ of true target samples is far closer to the center of source samples, compared to that of false target samples. Confining that the distribution of data representations between the source and target domain to be similar helps to reduce the influence caused by data noise to some extent.

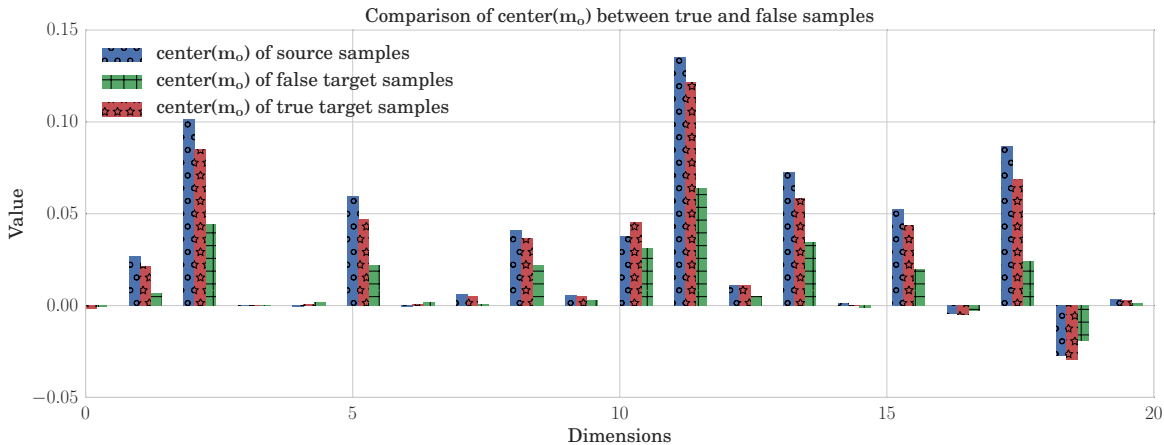

**Fig. 3.** Comparison of the center of $m_o$ between true and false samples on the first 20 dimensions. The center of $m_o$ of true target samples is far closer to the center of source samples, compared to that of false target samples. This observation supports our assumption that false instances among auto-annotated target samples tend to mutate the distribution of data representations on the element-wise multiply layer.

To encode this similarity, we utilize MMD (maximum mean discrepancy)[12] to compute distance between distributions of the element-wise multiply layers of the source domain and the target domain:

$$L_{EWM}(\theta^{T,n}|\mathbf{X}^S, \mathbf{X}^{T,n}, \theta^S) = \frac{1}{N^O} \sum_{o=1}^{N^O} \| \frac{\sum_{j=1}^{N^{T,n}} (\bm{m}_o^{T,n}|x_j^{T,n})}{N^{T,n}} - \frac{\sum_{i=1}^{N^S} (\bm{m}_o^S|x_i^S)}{N^S} \|^2 \quad (8)$$

which can also interpreted as the Euclidean distance between the center of $\bm{m}_o^{T,n}$ and $\bm{m}_o^S$ across all output dimensions. As a comparison, the MMD regularizer on feature vector layer can be formulated as:

$$L_{FV}(\theta^{T,n}|\mathbf{X}^S, \mathbf{X}^{T,n}, \theta^S) = \| \frac{\sum_{j=1}^{N^{T,n}} (\bm{f}^{T,n}|x_j^{T,n})}{N^{T,n}} - \frac{\sum_{i=1}^{N^S} (\bm{f}^S|x_i^S)}{N^S} \|^2 \quad (9)$$

where $\bm{f}$ is data of the feature vector layer in Equation 4 and $\bm{m}_o$ is the data of the element-wise multiply layer in Equation 6.

Since it's unpractical to get the distribution of the whole training set, while too few images cannot obtain a stable distribution for regularization. In our experiments, the $L_{EWM}(\cdot)$ loss is calculated for every batch. An example comparison of centers of $\bm{m}_o^S$ of different batches are shown in Fig 4.

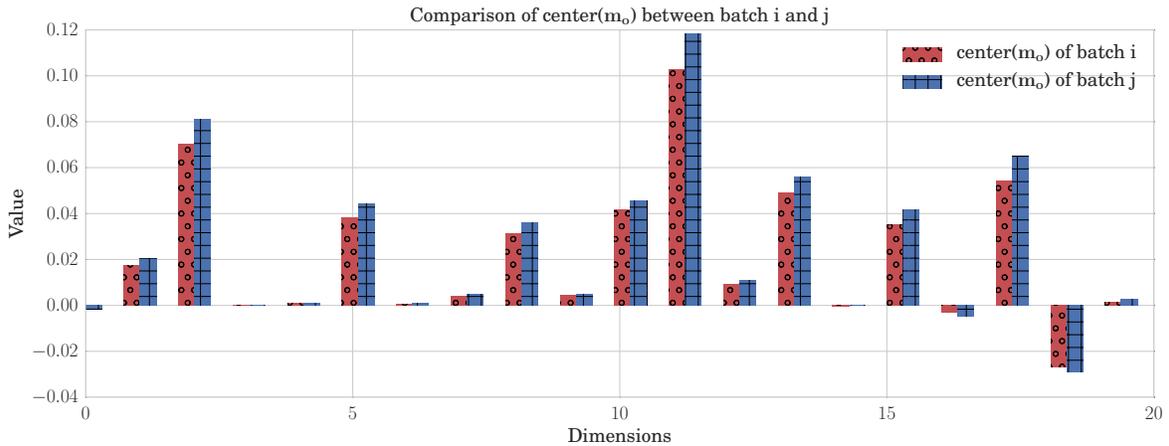

**Fig. 4.** Comparison of the center of $\bm{m}_o$ between two different batches on the first 20 dimensions. These two centers are close to each other, which supports our assumption that data distributions on the element-wise multiply layer between the source and target domain should be similar.

## 4   Experiment Results

In this section, we introduce our experiment results on both surveillance applications and the standard domain adaptation dataset. We firstly evaluate our approach on video surveillance. Then we employ our approach to standard domain adaptation benchmarks on both supervised and unsupervised settings to demonstrate the effectiveness of our method.

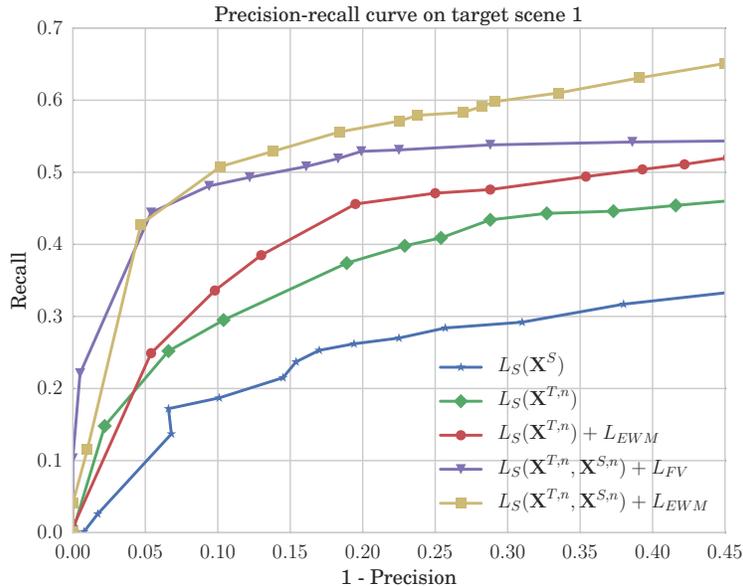

**Fig. 5.** Precision-recall curve of 5 comparison methods on target scene 1.

### 4.1   Domain Adaptation on Crowd Dataset

**Dataset and evaluation metrics** To show the effectiveness of our domain adaptation approach for pedestrian detection, we collected a dataset[1] consisting of 3 target scenes for the target domain. These three scenes contain 1308, 1213 and 331 unlabelled images, respectively. For each scene, 100 images are annotated for evaluation. Instead of labelling the whole body of a person, we label the head of a person as bounding box during training. The motivation for labelling only pedestrian heads comes from detection of indoor pedestrian or in crowded scenes, where the body of a person may be invisible. The dataset for the source domain are Brainwash Dataset[15].

Our evaluation metrics for detection uses the protocol defined in PASCAL VOC [16]. To judge a predicted bounding box whether correctly matches a ground truth bounding box, their intersection over their union must exceed 50%. And Multiple detections of the same ground truth bounding box are regarded as one correct prediction. For overall performance evaluation, the F1 score $F1 = 2*precision*recall/(precision+recall)$ [17] are utilized. Higher F1 score means better performance. At the same time, the precision-recall curves are also plotted.

**Experimental settings** Our generic detection model of adaptation architecture can be implemented by many deep detection models. In our experiment, we use the model proposed by Stewart et al. [15], which is an end to end detection network without any precomputed region proposals needed. For each iteration, 100 auto-annotated images from the target domain and 1000 annotated images from the source domain are alternatively used for training. The outputs of our detection network include bounding box locations and corresponding confidences, thus there are two fully connected layers between the last feature vector layer and the final outputs. In our experiments, when an unsupervised regularizer on the element-wise multiply layer predicting box confidence is added already, the unsupervised regularizer on the element-wise multiply layer predicting bounding box locations have little performance improvement. Experiments on 3 target scenes are executed separately.

---

[1] Our dataset will be made available on http://wylin2.drivehq.com/.

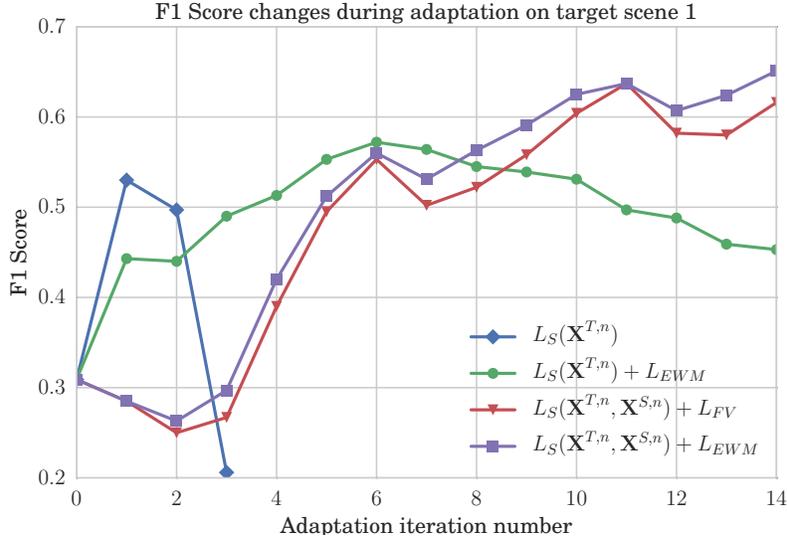

**Fig. 6.** F1 score changes of 5 comparison methods during adaptation on target scene 1

**Comparison with different methods** To demonstrate the effectiveness of our approach, 5 methods are compared among which method $L_S(\mathbf{X}^{T,n}, \mathbf{X}^{S,n}) + L_{EWM}$ is our final approach:

$L_S(\mathbf{X}^S)$ Source model only trained from the source domain.

$L_S(\mathbf{X}^{T,n})$ Only auto-labeled samples on the target domain are used for training, and without any unsupervised regularizer.

$L_S(\mathbf{X}^{T,n}) + L_{EWM}$ Only auto-labeled samples on the target domain are used for training, with an unsupervised MMD regularizer added on the last element-wise multiply layer.

$L_S(\mathbf{X}^{T,n}, \mathbf{X}^{S,n}) + L_{FV}$ Both auto-labeled images from the target domain and labeled images from the source domain are alternately sampled for training, with an unsupervised MMD regularizer [13] added on the last feature vector layer.

$L_S(\mathbf{X}^{T,n}, \mathbf{X}^{S,n}) + L_{EWM}$ Both auto-labeled images from the target domain and labeled images from the source domain are alternately sampled for training, with an unsupervised MMD regularizer added on the last element-wise multiply layer.

Fig 5 plots the precision-recall curves of the above comparison methods in target scene 1. Also, the changes of F1 score of every adaptation iteration are also depicted in Fig 6. Table 1 gives concrete precision and recall value of the 5 comparison methods on three target scenes when the F1 scores are at their highest. Examples of adaptation results are shown in Fig 7.

**Table 1.** Detection results of 5 compared methods on 3 target scenes

|  | Scene 1 | | | Scene 2 | | | Scene 3 | | |
|---|---|---|---|---|---|---|---|---|---|
|  | 1-Pr | Re | F1 | 1-Pr | Re | F1 | 1-Pr | Re | F1 |
| $L_S(\mathbf{X}^S)$ | 0.101 | 0.187 | 0.309 | 0.015 | 0.683 | 0.807 | 0.035 | 0.412 | 0.577 |
| $L_S(\mathbf{X}^{T,n})$ | 0.245 | 0.408 | 0.530 | 0.632 | 0.905 | 0.524 | 0.176 | 0.778 | 0.800 |
| $L_S(\mathbf{X}^{T,n}) + L_{EWM}$ | 0.284 | 0.476 | 0.572 | 0.012 | 0.837 | **0.906** | 0.078 | 0.653 | 0.764 |
| $L_S(\mathbf{X}^{T,n}, \mathbf{X}^{S,n}) + L_{FV}$ | 0.109 | 0.496 | 0.637 | 0.002 | 0.721 | 0.838 | 0.044 | 0.611 | 0.746 |
| $L_S(\mathbf{X}^{T,n}, \mathbf{X}^{S,n}) + L_{EWM}$ | 0.140 | 0.530 | **0.656** | 0.006 | 0.811 | 0.893 | 0.097 | 0.778 | **0.836** |

**Performance evaluation** From the Table 1, we have the following observations:

- Compared to method $L_S(\mathbf{X}^S)$, the recall values of other methods, which all utilize iterative algorithm for training, are explicitly larger. This implies the effectiveness of our iterative algorithm on boosting recall.
- The average F1 score of $L_S(\mathbf{X}^{T,n}) + L_{EWM}$ are larger than that of method $L_S(\mathbf{X}^{T,n})$. Also, the average (1-precision) value of $L_S(\mathbf{X}^{T,n}) + L_{EWM}$ is far smaller. Their difference in whether the unsupervised regularizer is added into loss function demonstrates that our unsupervised regularizer can mitigate the influence of data noise and thus boost F1 score.
- Compared to method $L_S(\mathbf{X}^{T,n}) + L_{EWM}$, the average F1 score of method $L_S(\mathbf{X}^{T,n}, \mathbf{X}^{S,n}) + L_{EWM}$ is higher. This demonstrate the effectiveness of negative source samples added into the training set during adaptation process.
- Compared to method $L_S(\mathbf{X}^{T,n}, \mathbf{X}^{S,n}) + L_{FV}$, the recall values of method $L_S(\mathbf{X}^{T,n}, \mathbf{X}^{S,n}) + L_{EWM}$ are further increased. This shows that unsupervised regularizer added on the element-wise layer will provide better regularizer effect compared to that on the feature vector layer.
- Our final method $L_S(\mathbf{X}^{T,n}, \mathbf{X}^{S,n}) + L_{EWM}$ achieves best results on target scene 1 and target scene 3. The performance on target scene 2 is rather close to the best result, which may result from large discrepancy of background between the source and target domain.

### 4.2   Domain Adaptation on Standard Classification Benchmark

In order to further demonstrate the effectiveness and generalization of our adaptation architecture, we test our method on the standard domain adaptation benchmark Office dataset[1].

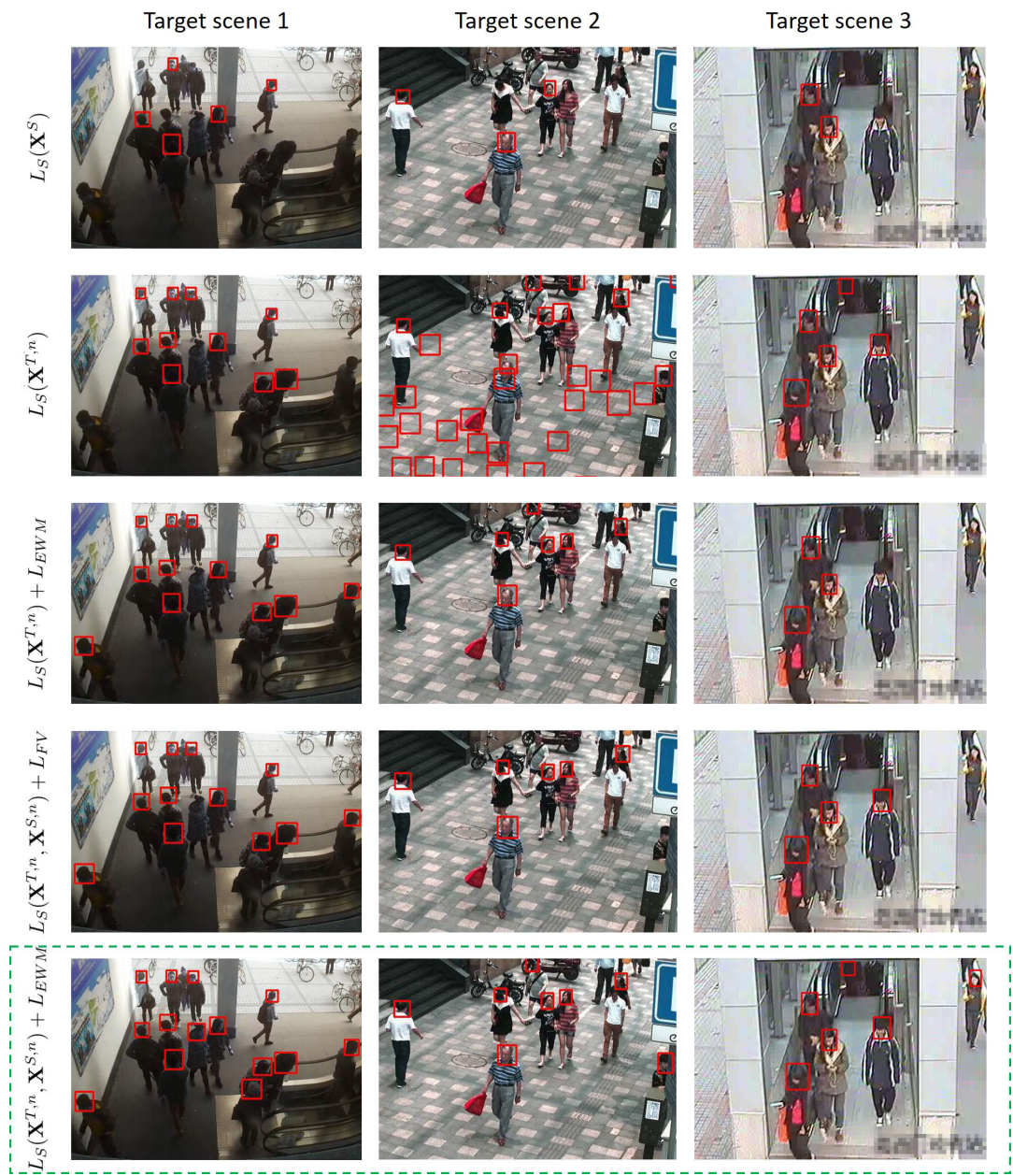

**Fig. 7.** Example results of 5 comparison methods on 3 target scenes.

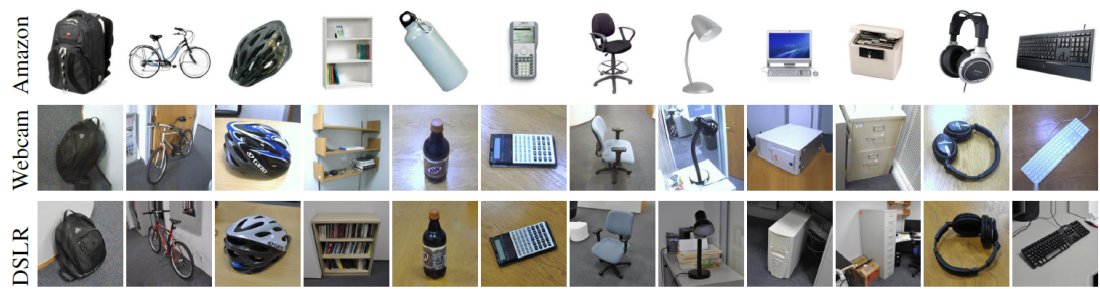

**Fig. 8.** Example images on Office dataset.

**Office dataset and Experimental settings** The Office dataset comprises 31 categories of objects from 3 domains (Amazon, DSLR, Webcam). Example images are depicted in Fig. 8. We take Amazon domain as the source domain and Webcam domain as the target domain. We follow the standard protocol for both supervised and unsupervised settings. We reused the architecture in pedestrian detection and utilize AlexNet [18] as the generic model of both streams.

**Performance evaluation** In Table 2, we compare our approach with other seven recently published works in both supervised and unsupervised settings. The outstanding performance on both settings confirms the effectiveness of our iterative algorithm and MMD regularizer on the element-wise multiply layer.

**Table 2.** Multi-class accuracy evaluation on Office dataset with supervised and unsupervised settings.

|  | A → W | |
|---|---|---|
|  | Supervised | Unsupervised |
| GFK(PLS,PCA)[19] | 46.4 | 15.0 |
| SA [20] | 45.0 | 15.3 |
| DA-NBNN [21] | 52.8 | 23.3 |
| DLID [22] | 51.9 | 26.1 |
| DeCAF$_6$S [23] | 80.7 | 52.2 |
| DaNN [11] | 53.6 | 35.0 |
| DDC[13] | 84.1 | 59.4 |
| Ours | **85.4** | **69.3** |

## 5  Conclusions

In this paper, we introduce an adaptation architecture to learn scene-specific deep detectors for the target domains. Firstly, an iterative algorithm is utilized to iteratively auto-annotate target samples and update the target model. As auto-annotated data are lack of negative samples and contain data noise, we randomly sample negative instances from the source domain. At the same time, an unsupervised regularizer is also designed to mitigate influence from data noise. More importantly, we propose to transform the last fully connected layer into an element-wise multiply layer and a sum layer for better regularizer effect.

## Acknowledgement

The work is partially funded by the following grants: DFG (German Research Foundation) YA 351/2-1, NSFC 61471235, Microsoft Research Asia Collaborative Research Award. The authors gratefully acknowledge the support.